\documentclass{article} % For LaTeX2e
\usepackage{iclr2015,times}
\usepackage{hyperref}
\usepackage{url}

\usepackage{times}
\usepackage{helvet}
\usepackage{courier}
\usepackage{amsmath}
\usepackage{amsfonts}
\usepackage{algorithm}
\usepackage{algorithmic}
\usepackage{graphicx,array,hyperref}
\usepackage{multirow}
\usepackage{hhline}
\usepackage{booktabs}
\usepackage{wrapfig}

\title{Restricted Boltzmann Machine for Classification with Hierarchical Correlated Prior}

\author{
Gang Chen \& Sargur N. Srihari %\thanks{ Use footnote for providing further information
\\
Department of Computer Science and Engineering\\
State University of New York at Buffalo\\
Buffalo, NY 14260, USA \\
\texttt{gangchen@buffalo.edu, srihari@cedar.buffalo.edu} \\
}

\iclrfinalcopy % Uncomment for camera-ready version

\begin{document}

\maketitle

\begin{abstract}
Restricted Boltzmann machines (RBM) and its variants have been widely used on classification problems. In a sense, its success of RBM  should be attributed to its strong representation power with hidden variables. Often, classification RBM ignores the interclass relationship or prior knowledge of sharing information among classes. %In this paper, we are interested in RBM with the hierarchical prior over classes, where parameters for nearby nodes are correlated in the hierarchical tree, and further the parameters at each node of the tree to be orthogonal to those at its ancestors. 
In this paper, we propose a RBM with hierarchical prior for classification problem, by generalizing the classification RBM with sharing information among different classes.  Basically, we assume the hierarchical prior over classes, where parameters for nearby nodes are correlated in the hierarchical tree, and further the parameters at each node of the tree to be orthogonal to those at its ancestors. Through the hierarchical prior, our model improves the information sharing between different classes and reduce the redundancy for robust classification. We test our method on several datasets, and show promising results compared to competitive baselines. 

%are interested in RBM for classification problems with prior assumption, where the sets of labels are expressed in a hierarchy. In order to reduce the redundancy between node parameters in the hierarchy, we also introduce orthogonal restrictions to our objective function. We test our method on challenge datasets, and show promising results compared to competitive baselines. 

%contain at least 70 and at most 150 words. It should be written using the
%\emph{abstract} environment.
%\keywords{restricted Boltzmann machines, hierarchical prior, hierarchical classification}
\end{abstract}

% -------------------------------------------------------------------------------------
\section{Introduction}
%Boltzmann machines are a type of neural networks that represent probability distributions in the form of scalar energy associated with each configuration of its variables. 
Restricted Boltzmann machines (RBM) \cite{Hinton02} are a specific neural network with no hidden-hidden and visible-visible connections. They have attracted significant attention recently on many machine learning problems, such as dimension reduction \cite{Hinton06b}, text categorization \cite{Larochelle12}, collaborative filtering \cite{Salakhutdinov07} and object recognition \cite{Krizhevsky12}. A recent survey \cite{Bengio12} shows how to improve classification accuracy by exploiting prior knowledge about the world around us. The purpose of this paper is to answer whether we can leverage the hierarchical structure over categories to improve the classification accuracy. %semantiwith prior knowledge of sharing information among classes in a hierarchy. 
The hierarchical tree \cite{Fellbaum98,Weigend99,Goodman01,Dekel04} over different classes is an efficient and effective way for knowledge representation and categorization. The top level of the taxonomy hierarchies starts with a general or abstract description of common properties for all objects, while the low levers add more specific characteristics. In other words, the semantic relationship among classes  is constructed from generalization to specification as depth increasing in the hierarchical tree or taxonomy. 
%Organizing different classes in a hierarchy \cite{Fellbaum98,Weigend99,Goodman01,Dekel04} is an efficient and effective way for knowledge representation and categorization. 
For example, WordNet \cite {Fellbaum98} and ImageNet \cite{Deng09} use this semantic hierarchy to model human psycholinguistic knowledge and object taxonomy respectively. %, refer to Fig. \ref{fig:taxo} for more information. %is a small part of hierarchical relationship described on WordNet \cite {Fellbaum98}. This can easily express ``cat" and ``dog" as carnivore, while ``cow" and ``sheep" as bovid, and all of them are placental animals.
Unfortunately, traditional RBM \cite{Larochelle08,Larochelle12} treats the category structure as flat and little work has been done to explore the interclass relationship. %We are interested in RBM for classification problems with prior assumption, where the sets of labels are expressed in a hierarchy. 
In this paper, we generalize RBM with hierarchical prior for classification problems. Basically, we divide the classification RBM into traditional RBM for representation learning and multinomial logit model for classification, see Fig. \ref{fig:crbm}(a) for intuitive understanding. For the traditional RBM (red in Fig. \ref{fig:crbm}(a)), we can extend it into deep belief network (DBN), while for the multinomial logit model (green in Fig. \ref{fig:crbm}(a)), we can incorporate the interclass relationship to it. In this work, we focus on the hierarchical prior over the classification RBM, and we take a similar strategy as corrMNL, that means we use sums of parameters along paths from root to a specific leaf in the tree as model parameters for hierarchical classification. However, we consider it in a rather different way from the previous work. We can think our method is a kind of mixture of corrMNL \cite{Shahbaba2007} and the orthogonal SVM model \cite{Xiao11}. However, our model inherits the advantage of RBM, which can learn the hidden representation for better classification \cite{Hinton06b,Larochelle12}, compared to the multinomial logit \cite{Shahbaba2007} and hierarchical SVM \cite{Dekel04,Xiao11}. Moreover, we only have a single RBM in our model, while there are multiple SVMs in the orthogonal hierarchical SVM \cite{Xiao11}.

%To the best of our knowledge, no work until now has incorporated hierarchical prior into RBM framework. 
Our contributions are: (1) we introduce the hierarchical semantic prior over labels into restricted Boltzmann machine; (2) we add orthogonal constraints over adjacent layers in the hierarchy, which makes our model more robust for classification problems. %hierarchical multi-class logistic regression. 
%More specifically, the regularization term forces the parameters of each node to be orthogonal to their parents as much as possible. %we require the regression coefficients for each leaf node is represented as the sums of parameters along paths to that leaf node in the tree. Meanwhile, we introduce regularization which forces the parameters of each node should be orthogonal to their parents as much as possible. 
%Thus, our model can effectively reduce the redundancy of parameters between children and parent, and also handle the hardness of classifying the similar classes at the leaf level. 
We test our method in the experiments, and show comparative results over competitive baselines. 
 
% -------------------------------------------------------------------------------
\section{Classification restricted Boltzmann machine with hierarchical correlated prior}\label{hcrbm}
We will revisit the classification RBM, then we will introduce our model. Throughout the paper, matrix variables are denoted with bold uppercases, and vector quantities are written in bold lowercase. For matrix ${\bf W}$, we indicate its $i$-th row and $j$-th column element as $W_{ij}$, its $i$-th row vector $W_{i.}$ and $j$-th column vector $W_{.j}$. For different matrixes, we use different subscripts to discern them. For example, ${\bf A}_{12}$ and ${\bf A}_{21}$ are different matrixes, which are indicated by different subscripts. 
\subsection{Classification Restricted Boltzmann Machine}\label{rbm}
%Restricted Boltzmann Machines (RBM) \cite{Hinton06b} are a particular form of Markov random field (undirected generative model), which are constructed with hidden nodes and visible nodes, and each connection in an RBM must link between a visible node and a hidden node (a bipartite graph, neither connection among visible nodes nor connection among hidden nodes). %use a layer of hidden variables to model a distribution over visible variables.
Denote $\mathcal{X} \in \mathbb{R}^d$ be an instance domain and $\mathcal{Y}$ be a set of labels. Assume that we have a training set $\mathcal{D} = \{({\bf x}_{i}, y_i)\}$, comprising for the $i$-th pair: an input vector ${\bf x}_{i} \in \mathcal{X}$ and a target class $y_{i} \in \mathcal{Y}$, where ${\bf x}_{i} \in \mathbb{R}^d$ and $y_{i} \in \{1,..., K\}$. An RBM with $n$ hidden units is a parametric model of the joint distribution between a layer of hidden variables ${\bf h} = (h_1,...,h_n)$ and the observations ${\bf x} = (x_1, ..., x_d)$ and $y$. 

The classification RBM was first proposed in \cite{Hinton07} and was further developed in \cite{Larochelle08,Larochelle12} with discriminative training model. %Basically, It \cite{Larochelle12} generalizes traditional RBM \cite{Hinton06b} which is undirected graphic model with a layer of hidden variables to model a distribution over visible variables, into the classification RBM with target class observations. To better understand this work, we think the classification RBM is composed of traditional RBM for feature learning and multinomial logistic regression classifier, see Fig. \ref{fig:crbm}(a) for explanation. 
The joint likelihood of the classification RBM takes the following form:
\begin{equation}\label{eq:eq1}
p(y,{\bf x},{\bf h}) \propto e^{-E(y,{\bf x},{\bf h})}
\end{equation}
where the energy function is 
\begin{equation}\label{eq:eq2}
E(y,{\bf x},{\bf h}) = -{\bf h}^T{\bf W}{\bf x} - {\bf b}^T {\bf x} - {\bf c}^T{\bf h} - {\bf d}^T {\bf y} - {\bf h}^T{\bf U} {\bf y}
\end{equation}
with parameters $\Theta = \{{\bf W}, {\bf b}, {\bf c}, {\bf d}, {\bf U}\}$ and ${\bf y} = (1_{y=i})_{i=1}^K$ for $K$ classes, where matrix ${\bf W} \in \mathbb{R}^{n \times d}$, and ${\bf U} \in \mathbb{R}^{n \times K}$.
%Further we can compute the following conditional likelihood:
%\begin{subequations}\label{eq:eq3}
%\begin{align}
%        & p({\bf h}|y,{\bf x} ) = \prod_{j} p(h_j|y, {\bf x})\\
%        & p({\bf x} | {\bf h}) = \prod_{i} p(x_{i} | {\bf h}) \label{eq:eq31} \\
%        & p(x_i=1 | {\bf h}) = \mathrm{logistic}(b_{i} + \sum_{j} W_{ij}h_j) \label{eq:eq32}  \\
%        & p(h_i=1 | {\bf v}) = \mathrm{logistic}(c_{i} + \sum_{j} W_{ji}v_j) \label{eq:eq321}  \\
%        & p(y | {\bf h}) = \frac{e^{d_{y}+\sum_{j} U_{jy}h_{j}} } {\sum_{y^*} e^{d_{y*}+\sum_{j} U_{jy*}h_{j}} } \label{eq:eq33} 
%\end{align}
%\end{subequations}
%where $\mathrm{logistic}(x)  = 1/(1+e^{-x})$ in Eq. (\ref{eq:eq32}). The prediction $p(y | {\bf h})$ given hidden variables ${\bf h}$ in Eq. (\ref{eq:eq33}) is the multinomial logit model (a.k.a multiclass logistic regression or softmax function). In other words, for any new input $x$, we can encode it into hidden space, and then use $p(y | {\bf h})$ for its label prediction. The green area in Fig. \ref{fig:crbm}(a) shows the prediction with softmax function.  

\begin{figure*}[t!]
\centering
\includegraphics[trim = 3mm 52mm 16mm 22mm, clip, width=12cm]{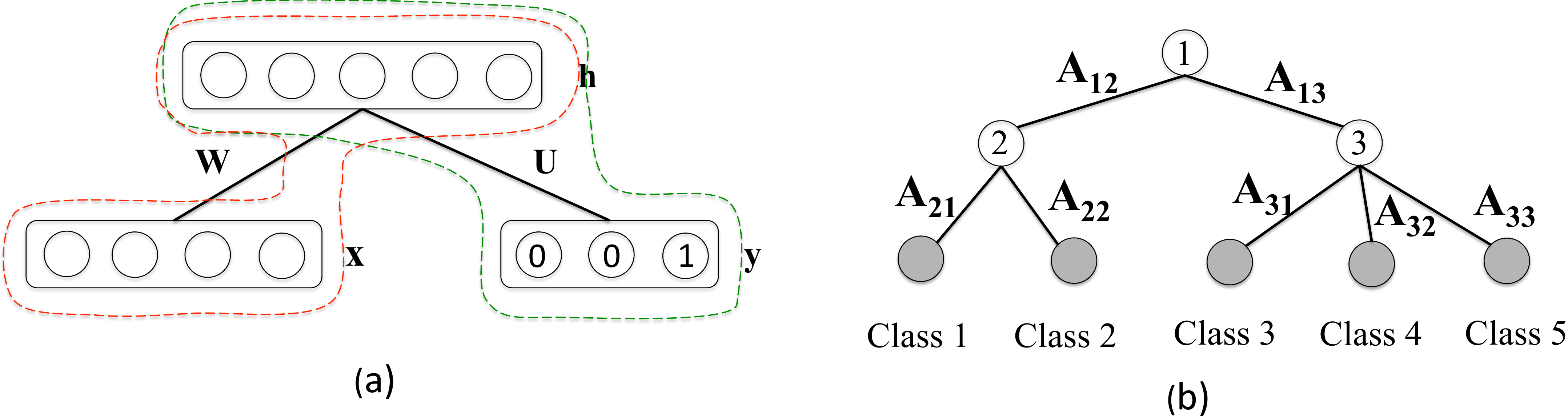}
\caption{(a) It is the classification restricted Boltzmann machine, which integrates restricted Boltzmann machine and logistic regression model; the left red dash area is restricted Boltzmann machine for representation learning, while the green region shows the logistic regression model for multi-class problems. (b) A hierarchical example for explanation, in which all internal nodes are depicted with white background, while leafs/classes are shown in gray in the hierarchy. The parameters for each classes are presented as a sum of parameters along its ancestors at different level of hierarchy. For example, the coefficient parameter of class 1 is ${\bf A}_{12} + {\bf A}_{21}$.}
\label{fig:crbm}
\end{figure*}

For classification problem, we need to compute the conditional probability for $p(y| {\bf x})$. As shown in \cite{Salakhutdinov07}, this conditional distribution has explicit formula and can be calculated exactly, by writing it as follows:
\begin{equation}\label{eq:eq4}
p(y | {\bf x})  = \frac{e^{d_{y}} \prod_{j=1}^n \big(1+e^{c_{j}+U_{jy} + \sum_{i} W_{ij}x_i} \big)}{\sum_{y^*} e^{d_{y^*}} \prod_{j=1}^n \big(1+e^{c_{j}+U_{jy^*} + \sum_{i} W_{ij}x_i} \big)}
\end{equation}

To learn RBM parameters, we need to optimize the joint likelihood $p(y, {\bf x})$ on training data $\mathcal{D}$. Note that it is intractable to compute $p(y, {\bf x})$, because it needs to model $p(x)$. Fortunately,  Hinton proposed an efficient stochastic descent method, namely contrastive divergence (CD) \cite{Hinton02} to maximize the joint likelihood. Thus, we get the following stochastic gradient updates for ${\bf W}$ and ${\bf U}$ from CD respectively
\begin{equation}\label{eq:grad}
\frac{\partial{\mathrm{log}p({\bf x}, y)}}{\partial{W_{ij}}} = \langle v_{i}h_{j}\rangle_{data} -  \langle v_{i}h_{j} \rangle_{model}   \,\,\,\,\,\,\,\,\,
\frac{\partial{\mathrm{log}p({\bf x}, y)}}{\partial{U_{jk}}} = \langle h_{j}y_{k}\rangle_{data} -  \langle h_{j}y_{k} \rangle_{model} 
\end{equation}

And update $\Theta$ until convergence with gradient descent
\begin{equation}
\Theta = \Theta + \eta \frac{\partial \mathrm{log} p({\bf x}, y)}{\partial \Theta}
%\Theta = \Theta - \eta \frac{\partial \mathrm{log} \mathcal{L}_{dis}(\mathcal{D};\Theta)}{\partial \Theta})
\label{eq:grbm}
\end{equation}
where $\eta$ is the learning rate for the classification RBM. 

\subsection{Restricted Boltzmann machine with hierarchical prior}\label{hrbm}
Our model introduces hierarchical prior over label sets for logistic regression classifier in the classification RBM. Note that we divide the classification RBM into two parts: RBM (feature learning) and multinomial logit model (classifier), corresponding to red and green regions shown in Fig. \ref{fig:crbm}(a) respectively. Our model introduces the hierarchical prior over multinomial logit regression classifier, which is vital for classification problems under RBM framework. 

%Define $\mathcal{X} \in \mathbb{R}^n$ be an instance domain and $\mathcal{Y}$ be a set of labels in the hierarchical tree $\mathcal{T}$. 
Define the hierarchical tree $\mathcal{T} = (\mathcal{V}, \mathcal{E})$, the number of node $N = |\mathcal{V}|$ and the number of edge $M= |\mathcal{E}|$. Furthermore, we assume all parameters along edges are ${\bf A} = \{{\bf A}_1,.., {\bf A}_m\}$, where $\{{\bf A}_j\}_{j=1}^m$ describes the parameter for each edge in the hierarchy respectively and ${\bf A}_j$ has the same size as ${\bf U}$ in the above subsection \ref{rbm}.
%We denote $K = |\mathcal{Y}|$ as the number of classes, 
For any node $\nu$ in the tree, we denote $\mathcal{A}(\nu)$ as its direct parent (vertex adjacent to $v$), and $\mathcal{A}^{(i)}(\nu)$ to be its $i$-th ancestor of $\nu$. As in \cite{Dekel04}, we also define the path for each node $\nu \in \mathcal{T}$, define $P(\nu)$ to be the set of nodes along the path from root to $v$,
\begin{equation}
 P(\nu) = \{ \mu \in \mathcal{T}:  \exists i \quad \mu = \mathcal{A}^{(i)} (\nu)  \}
\end{equation}
Now we can define the coefficient parameters for each leaf node $\nu$ as 
\begin{equation}\label{eq:phi}
 \boldsymbol{\mathcal{A}}(\nu)= \sum_{\mu \in P(\nu)} {\bf A}_{\mu}
 \end{equation}
 where the classification coefficient for each class in Eq. (\ref{eq:phi}) is decomposed into contributions along paths from root to the leaf associated to that class. For our model, each leaf node is associated to one class, which takes the same methodology as in \cite{Salakhutdinov07}. Fig. \ref{fig:crbm}(b) is an example with total five classes, where the sums of parameters along the path to the leaf node are coefficient parameters used for classification. In Fig. \ref{fig:crbm}(b), ${\bf A}_{12}$ and ${\bf A}_{13}$ are parameters along branches in the first level, and ${\bf A}_{21}$, ${\bf A}_{22}$, ${\bf A}_{31}$, ${\bf A}_{32}$ and ${\bf A}_{33}$ are parameters in the second level. For example, the coefficient parameter of class 1 is ${\bf A}_{12} + {\bf A}_{21}$ according to Eq. (\ref{eq:phi}); similarly, for class 4, its coefficient parameter is ${\bf A}_{13} + {\bf A}_{32}$. For example, we can see class 1 and class 2 sharing the common term ${\bf A}_{12}$, which can be thought as the prior correlation between the parameters of nearby classes in the hierarchy.

 For $K$ classes, we have ${\bf U} \in \mathbb{R}^{n \times K}$ and ${\bf A}_{j} \in \mathbb{R}^{n \times K}$ for $j=\{1,..,m\}$. Thus we can factorize 
 \begin{equation}\label{eq:fact}
 {\bf U} = {\bf VA}
 \end{equation}
 where ${\bf A} = \{{\bf A}_1,.., {\bf A}_m\} \in \mathbb{R}^{mn \times K}$ is the concatenation of parameters $\{{\bf A}_j\}_{j=1}^m$ of all edges in the hierarchy, while ${\bf V} \in \mathbb{R}^{n \times mn}$ implies the hierarchical prior over labels, refer Eq. (\ref{eq:phi}) for construction of the correlated matrix ${\bf V}$. Note that ${\bf V}$ (just) encodes the given hierarchical structures with 0 or 1 and is fixed during training the models. In addition, we introduce orthogonal restrictions just as in \cite{Xiao11} to reduce redundancy between adjacent layers. Given a training set $\mathcal{D} = \{({\bf x}_{i}, y_i)\}$, we propose the following objective function:
\begin{equation}
\mathcal{L}(\mathcal{D}; \Theta) = - \sum_{i = 1}^{|D|} \mathrm{log} p(y_{i}, {\bf x}_{i}) + C\sum_{\nu,\mu \in P(\nu)} trace({\bf A}_{\mu}^T{\bf A}_{\nu})
\end{equation}
where $C$ is the weight to balance the two terms. The first term is from the negative log likelihood as in RBM and the second term forces parameters at children to be orthogonal to those at its ancestor as much as possible.

The differences between our model and RBM lie: (1) hierarchical prior over labels, which can induce correlation between the parameters of nearby nodes in the tree; (2) we have orthogonal regularization which can make our model more robust, and also reduce redundancy in model parameters. %Considering the constraints introduced here are only related to ${\bf U}$, our model have the same updates as RBM, excerpt ${\bf U}$.
For parameters updating, we have the same equations as in the classification RBM, except for ${\bf U}$ which introduces hierarchical prior and orthogonal restrictions among children-parent pairs. %Our method introduces the hierarchical correlation and also orthogonal regularization between parameters. 

According to chain rule, we can differenciate $\mathcal{L}(\mathcal{D}; \Theta)$ r.w.t ${\bf A}_{\nu}$ and get the following derivative
%\begin{subequations}\label{eq:chaingrad}
%\begin{align}
%%\frac{\mathcal{L}(\mathcal{D}; \Theta)}{\partial{A_{jk}}} =- \frac{\partial{\mathrm{log}p({\bf x}, y)}}{\partial{U_{jk}}} \cdot \frac{\partial{U_{jk}}}{\partial{A_{jk}}} +
%%\langle h_{j}y_{k}\rangle_{data} -  \langle h_{j}y_{k} \rangle_{model} 
%%\frac{\mathcal{L}(\mathcal{D}; \Theta)}{\partial{A_{jk}}} = \frac{\mathcal{L}(\mathcal{D}; \Theta)}{\partial{U_{jk}}} \cdot \frac{\partial{U_{jk}}}{\partial{A_{jk}}} 
%\frac{\mathcal{L}(\mathcal{D}; \Theta)}{\partial{{\bf A}_{\nu}}} = \frac{\mathcal{L}(\mathcal{D}; \Theta)}{\partial{{\bf U}}} \cdot \frac{\partial{{\bf U}}}{\partial{{\bf A}_{\nu}}}  + C{\bf A}_{\mu}
%\label{eq:hgrad3}
%\end{align}
%\end{subequations}
\begin{equation}\label{eq:chaingrad}
\frac{ \partial{  \mathcal{L}(\mathcal{D}; \Theta)} }{\partial{{\bf A}_{\nu}}} = -\frac{\partial{  \sum_{i = 1}^{|D|} \mathrm{log} p(y_{i}, {\bf x}_{i}) } }{\partial{{\bf U}}} \cdot \frac{\partial{{\bf U}}}{\partial{{\bf A}_{\nu}}}  + C \sum_{\mu \in P(\nu)}{\bf A}_{\mu} % C{\bf A}_{\mu} + 
\end{equation}
Note that the derivative of $\sum_{i = 1}^{|D|} \mathrm{log} p(y_{i}, {\bf x}_{i})$ w.r.t. ${\bf U}$ can be computed via Eq. (\ref{eq:grad}). Thus, we can use Eq. (\ref{eq:chaingrad}) to calculate derivative w.r.t. ${\bf A}_{\nu}$, and then update ${\bf A}_{\nu}$ with stochastic gradient descent. Given ${\bf A}_{\nu}$, we can use Eq. (\ref{eq:fact}) to update ${\bf U}$. 

\subsection{Algorithm}
Note that our model incorporates the hierarchical prior and orthogonal constraints through ${\bf U}$. In other words, we can update all parameters with CD, except ${\bf U}$. Because ${\bf U}$ is the function of ${\bf A}$, we can compute the derivative of ${\bf U}$ w.r.t. ${\bf A}$ and update ${\bf A}$ with gradient descent. After we get ${\bf A}$, we can calculate ${\bf U}$, which can be used in the next iteration. We list the pseudo code below in Alg. \ref{alg:alg1}.% and implement it in Matlab. 
\begin{algorithm}[h!]
  \footnotesize
\caption{Learning RBM with hierarchical correlated prior} 
\label{alg1}
\textbf{Input}: training data $\mathcal{D} = \{({\bf x}_{i}, y_i)\}$, the number of hidden nodes $n$, learning rate $\eta$, $C$ and maximum epoch $T$ \\%and $\epsilon$.
\textbf{Output}: $\Theta = \{{\bf W}, {\bf b}, {\bf c}, {\bf d}, {\bf U}\}$
\begin{algorithmic}[1]
\STATE Initialize parameters ${\bf W}, {\bf b}, {\bf c}, {\bf d}, {\bf U}$;
\STATE Divide the training data into batches;
\FOR{$t=1$ to $T$}
\FOR{each batch}
\STATE Use 1-step Gibbs sampling to update the gradient according to Eq. (\ref{eq:grad});
\STATE Update all other parameters except ${\bf U}$ with CD; 
%\COMMENT{the following steps will update ${\bf U}$} 
\STATE Compute gradient w.r.t. ${\bf A}_{\nu}$ according to Eq. (\ref{eq:chaingrad});
\STATE Update ${\bf A}$ with gradient descent with Eq. (\ref{eq:grbm});
\STATE Update ${\bf U}$ according to Eq. (\ref{eq:fact});
\ENDFOR 
\ENDFOR
\STATE Output ${\bf W}, {\bf b}, {\bf c}, {\bf d}, {\bf U}$;
\STATE End
\end{algorithmic}
\label{alg:alg1}
\end{algorithm}

\begin{figure*}[t!]
\centering
\begin{tabular}{cc}
\includegraphics[trim = 15mm 26mm 10mm 19mm, clip, height = 2.7cm, width=5.5cm]{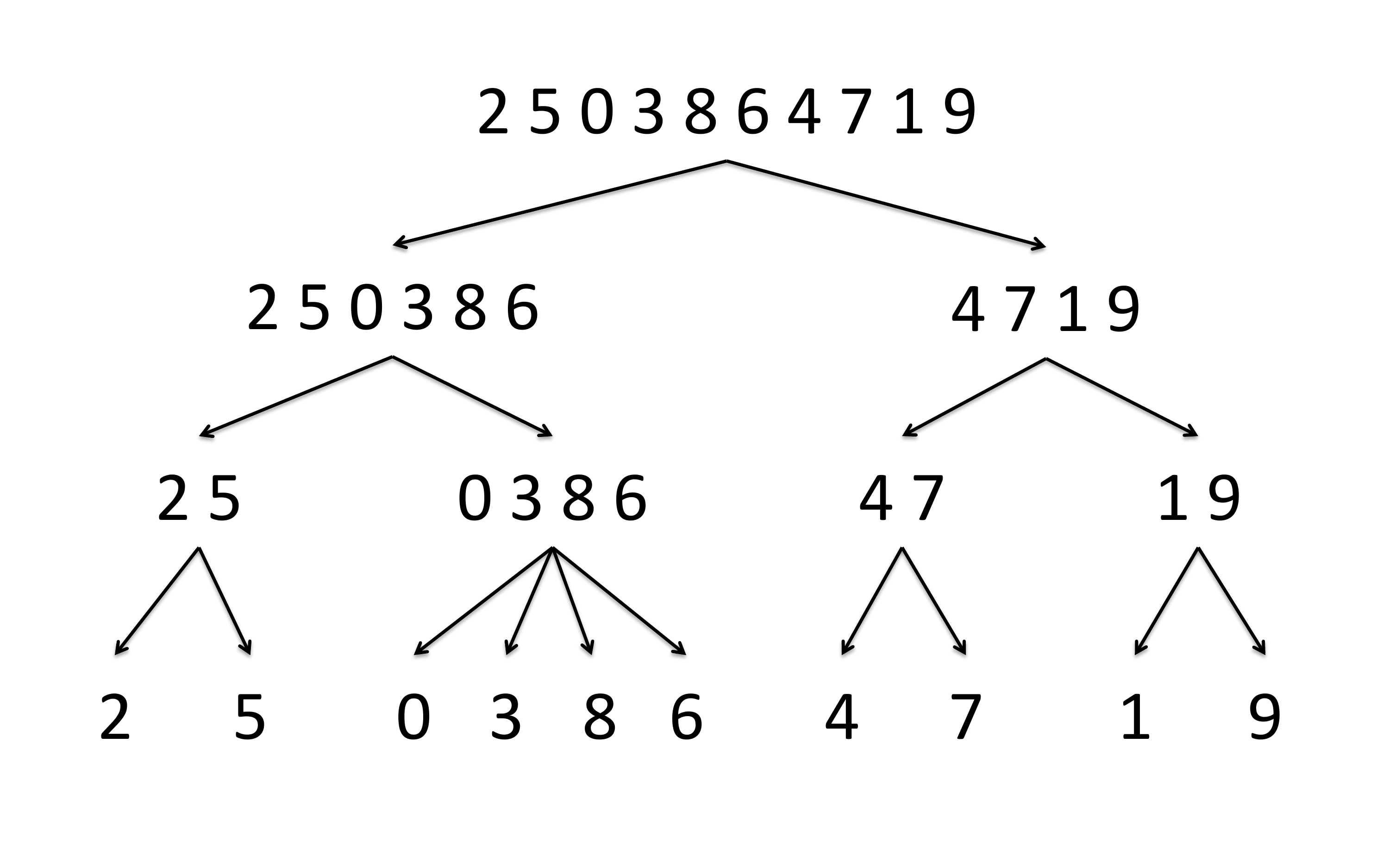} & 
\includegraphics[trim = 5mm 26mm 15mm 19mm, clip, height = 2.7cm, width=5.5cm]{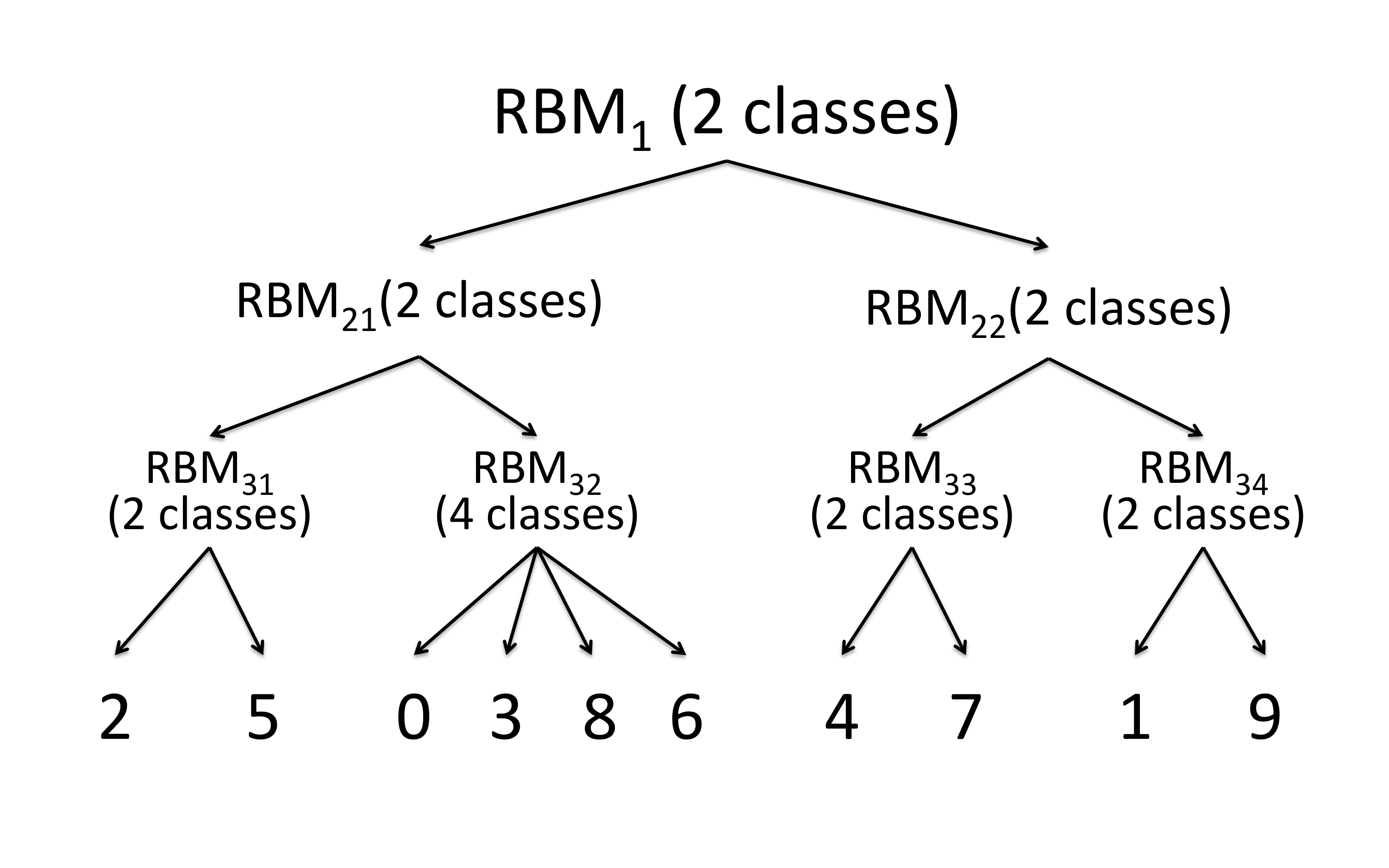} \\
(a) & (b)
\end{tabular}
\caption{(a) The hierarchical structure prior over label sets from MNIST digital dataset; we use this prior over labels with the purpose to capture similar structure information between different characters. For example, `3' and `8' share some parts, and similar structure information can be found in pairs `4'  and `7', as well as  `1' and `9'. (b) The hierarchical classification RBM (HRBM), which is constructed according the hierarchical prior Fig. \ref{fig:hrbmprior}(a). In order to learn HRBM classifier, we learn a RBM classifier for each node and recursively to the leafs in a top-down manner.}%(b) The hierarchical classification RBM (HRBM). HRBM is constructed according the hierarchical prior (left side graph). In order to learn HRBM classifier, we learn a RBM classifier for each node and recursively to the leafs in a top-down manner. }
\label{fig:hrbmprior}
\end{figure*}

\begin{figure*}[t!]
\centering
\begin{tabular}{c} 
\includegraphics[trim = 3mm 1mm 1mm 2mm, clip, height = 3.2cm, width=14cm]{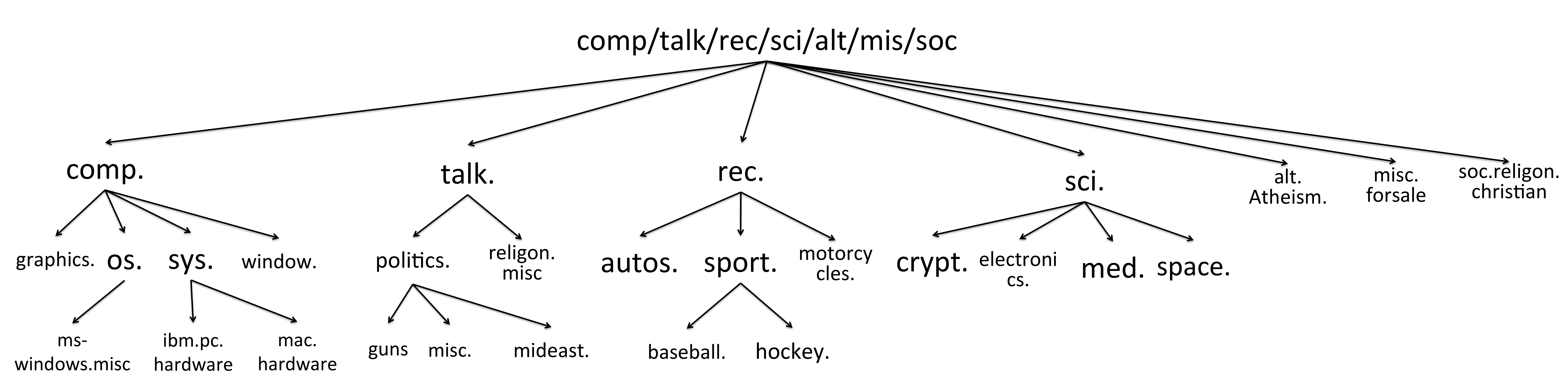} 
\end{tabular}
\caption{The hierarchical structure from 20 news group dataset. The root (or the first level) cover documents from all categories, while the leaf level indicates labels where documents attached to.}%(b) The hierarchical classification RBM (HRBM). HRBM is constructed according the hierarchical prior (left side graph). In order to learn HRBM classifier, we learn a RBM classifier for each node and recursively to the leafs in a top-down manner. }
\label{fig:hrbm}
\end{figure*}

% ----------------------------------------------------------------------------------------------
\section{Experimental Results}
We analyze our model with experiments on two classification problems: character recognition and document classification, and compare our results to those from competitive baselines below. 
\\
%\begin{description}
{\bf RBM}  or RBM for classification was first proposed in \cite{Hinton06b} and later was further developed in \cite{Larochelle12}. Its mathematical formula is shown in Eq. (\ref{eq:eq2}). \\
%\item[Hierarchical classification RBM] with soft assignment (HRBMs) \hfill \\
{\bf Hierarchical classification RBM} with soft assignment (HRBMs) is a nested hierarchical classifier in a top-down way, shown in Fig. \ref{fig:hrbmprior}(b). In the training stage, for each internal node (including root node) in the current level, HRBM will split training data according to its children and learn a classification RBM for multiple classes (decided by the number of its children). In the inference stage, the likelihood for certain classes in the current layer depends both on the output probability of this layer classifier and also the conditional likelihood on the upper levels. For example, the probability to assign label 2 to a given instance in Fig. \ref{fig:hrbmprior}(b) depends on the output probabilities from $\textrm{RBM}_1$, $\textrm{RBM}_{21}$ and $\textrm{RBM}_{31}$. %HRBMs computes the classification probabilities for each node in each level, until to leaf nodes. 
For each data instance, its probability belongs to each class is the probability production along path from root to the leaf of that class, and finally we assign the data instance to the label with largest probability. 
%\item[Hierarchical classification RBM] with hard assignment (HRBMh)\hfill \\
\\
{\bf Hierarchical classification RBM} with hard assignment (HRBMh) has the similar hierarchical structure as HRBMs in Fig. \ref{fig:hrbmprior}(b).
%The HRBMh learns RBM classifiers in each level and recursively classify data in a top-down way, shown in Fig. \ref{fig:hrbmprior}(b).  
%More specifically, HRBMh is a special case of HRBM, which assigns labels according to output probabilities in each level until to leafs, instead of assigning probability (soft assignment) as HRBMs in each hierarchical level. 
The difference between HRBMs and HRBMh is that HRBMs assign classification probability to each node, while HRBMh assign labels.  
\\
%\item[Hidden hierarchical classification RBM] (HHRBM) \hfill \\
{\bf Hidden hierarchical classification RBM} (HHRBM) is similar as the hierarchical classification RBM (HRBM) in a top-down manner. For any current node, HHRBM learns a classification RBM and projects the training data into hidden space for its children (Note that RBM can map any input instance into its hidden space). Then, all its children recursively learn classification RBMs with projected hidden states as input from its parent node until to leaf level. In a sense, HHRBM works similar to the deep believe network (DBN) in \cite{Hinton07}. Hence, the only difference between HHRBM and HRBM is that HRBM computes the classification probability with the visual data as input for all levels, while HHRBM calculates the classification probability with hidden states as input in a top-down manner. 
\\
%\item[Multinomial logit model] (MNL) \hfill \\
{\bf Multinomial logit model} (MNL), a.k.a multiclass logistic regression, has no class correlated hierarchical structure. 
\\
%\item[Correlated Multinomial logit regression] (corrMNL) \hfill \\
{\bf Correlated Multinomial logit regression} (corrMNL) \footnote{\url{http://www.ics.uci.edu/~babaks/Site/Codes.html}} extends MNL with hierarchical prior over classes, refer to \cite{Shahbaba2007} for more details. 
%\end{description}
  
In all the above baselines, HRBMs, HRBMh, HHRBM and corrMNL leverage the hierarchical prior over label sets for classification, while RBM and MNL have no such prior information available. As for the difference in the number of RBMs used, (H)HRBMs belong to the tow-down classification approaches where multiple RBMs are constructed and each of which is trained to classify training examples into one of its children in a hierarchical tree while our approach maintains only a single RBM. 

%\begin{figure}[h!]
%\centering
%\includegraphics[trim = 45mm 100mm 45mm 95mm, clip, width=8cm]{mnist_ex.pdf}
%\caption{Example images from MNIST dataset.}
%\label{fig:mnist}
%\end{figure}
\noindent{\bf Character Recognition} MNIST digits\footnote{\url{http://yann.lecun.com/exdb/mnist/}} consists of $28\times28$-size images of handwriting digits from $0$ through $9$%with a training set of 60,000 examples and a test set of 10,000 examples
, and has been widely used to test character recognition methods. %A set of examples are shown in Fig. (\ref{fig:mnist}). 
In the experiment, we use Fig. \ref{fig:hrbmprior}(a) as our hierarchical prior over label sets. To test our method and other baselines, we sample 5000 images from the training sets as our training examples and 1000 examples from the testing sets as our testing data. The reason that we use a subset of MNIST is to answer whether the correlation between different classes is valuable for classification problem when the number of training examples for individual classes may be relatively small. In order to make our method comparable to other baselines, we have the same parameter setting for RBM related methods (including RBM, HRBMs, HRBMh and our method). We set the number of hidden states $n =100$ and the learning rate $\eta = 0.1$ for RBM related methods, and the extra parameter in our model $C = 0.1$. Both HRBMs and HRBMh learn a RBM for each node and recursively to leafs, shown in Fig. \ref{fig:hrbm}. For the HHRBM with 4 layers decided by the hierarchical prior in Fig. \ref{fig:hrbmprior}(a), we set its number of hidden states 100, 50, 25 and 20 for each layer respectively.

\begin{table}[t!]
\centering
\resizebox{0.85\columnwidth}{!}{%
\begin{tabular}{llccccccc}
\hline
\multicolumn{9}{ c }{Error Rate (\%)}  \\
\hline\hline
%\multicolumn{2}{c}{Cursive} & Print \\
%\cline{1-2}
%Methods    & Patches & error rate (\%) \\
%\hline
\multirow{2}{*}{Datasets} & \multicolumn{8}{c}{Model} \\
\cline{2-9}
& SVM & MNL & corrMNL & HRBMh & HRBMs & HHRBM & RBM & Ours\\
\hline
MNIST & 10.8&10.6  & 8.97 & 12.1 & 7.95 & 11.10 &8.22  & {\bf 7.91}\\
%  \hline
%  20news & 30.8 & - & 30.6 & 0.026 & 32.2 & 24.9 & 23.9\\
\cline{1-9}
\end{tabular}}
\caption{The experimental comparison on a subset of MNIST dataset, with total 5000 training examples and 1000 testing samples. We compare the performances between our method and the baselines. It demonstrates that our method with hierarchical prior over labels can improve recognition accuracy. }
\label{tab:tab3}
\end{table}
The comparison between our method and the baselines is shown in Table (\ref{tab:tab3}). Our method incorporates the hierarchical prior structure over labels, and the experimental results show that our method outperform other RBM related methods, and also demonstrates that the hierarchical prior in our method is helpful to improve the recognition accuracy. 
\begin{figure*}[t!]
\centering
\begin{tabular}{cc}
\includegraphics[trim = 35mm 80mm 40mm 80mm, clip, width=6cm]{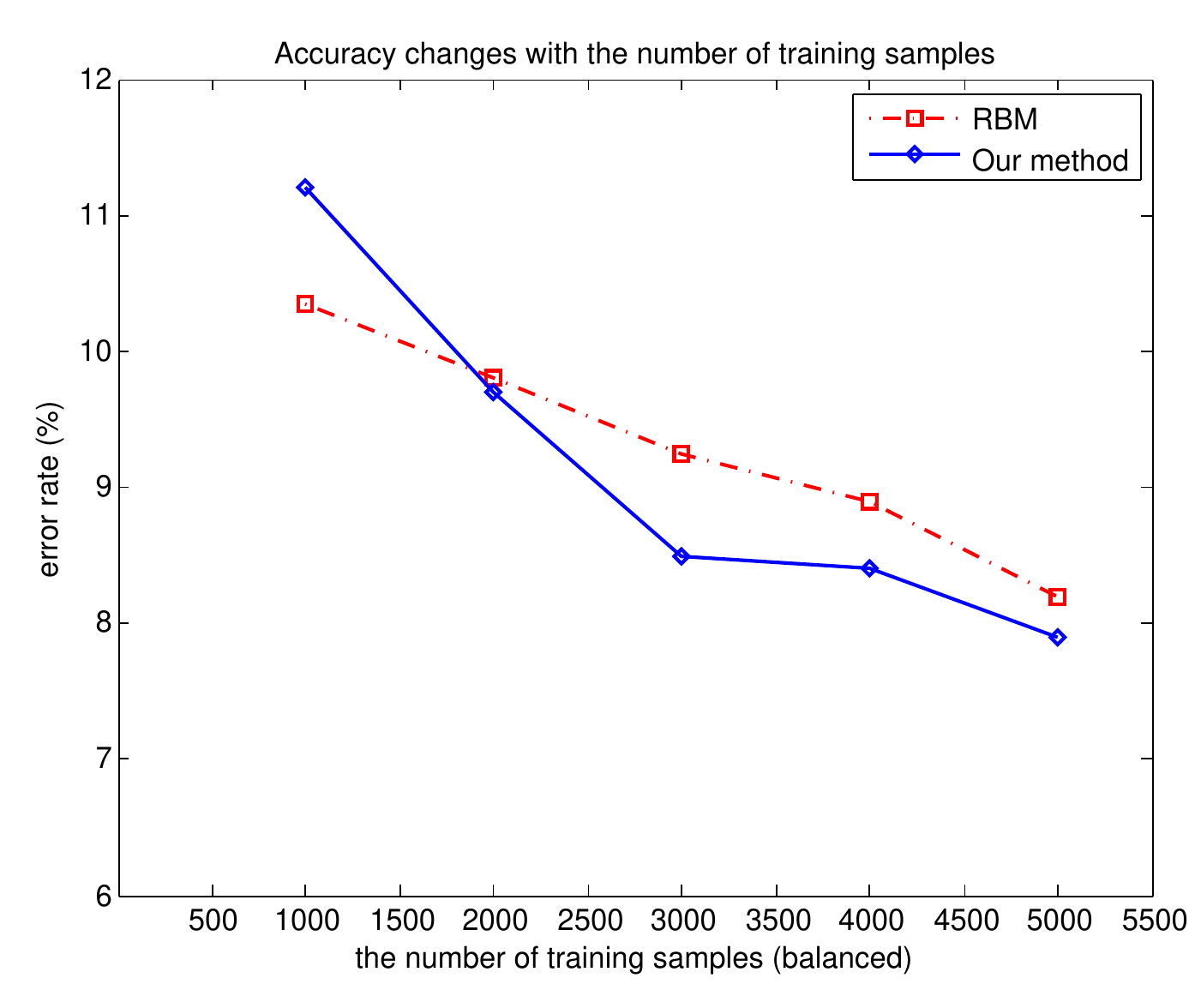} &
\includegraphics[trim = 35mm 80mm 40mm 80mm, clip, width=6cm]{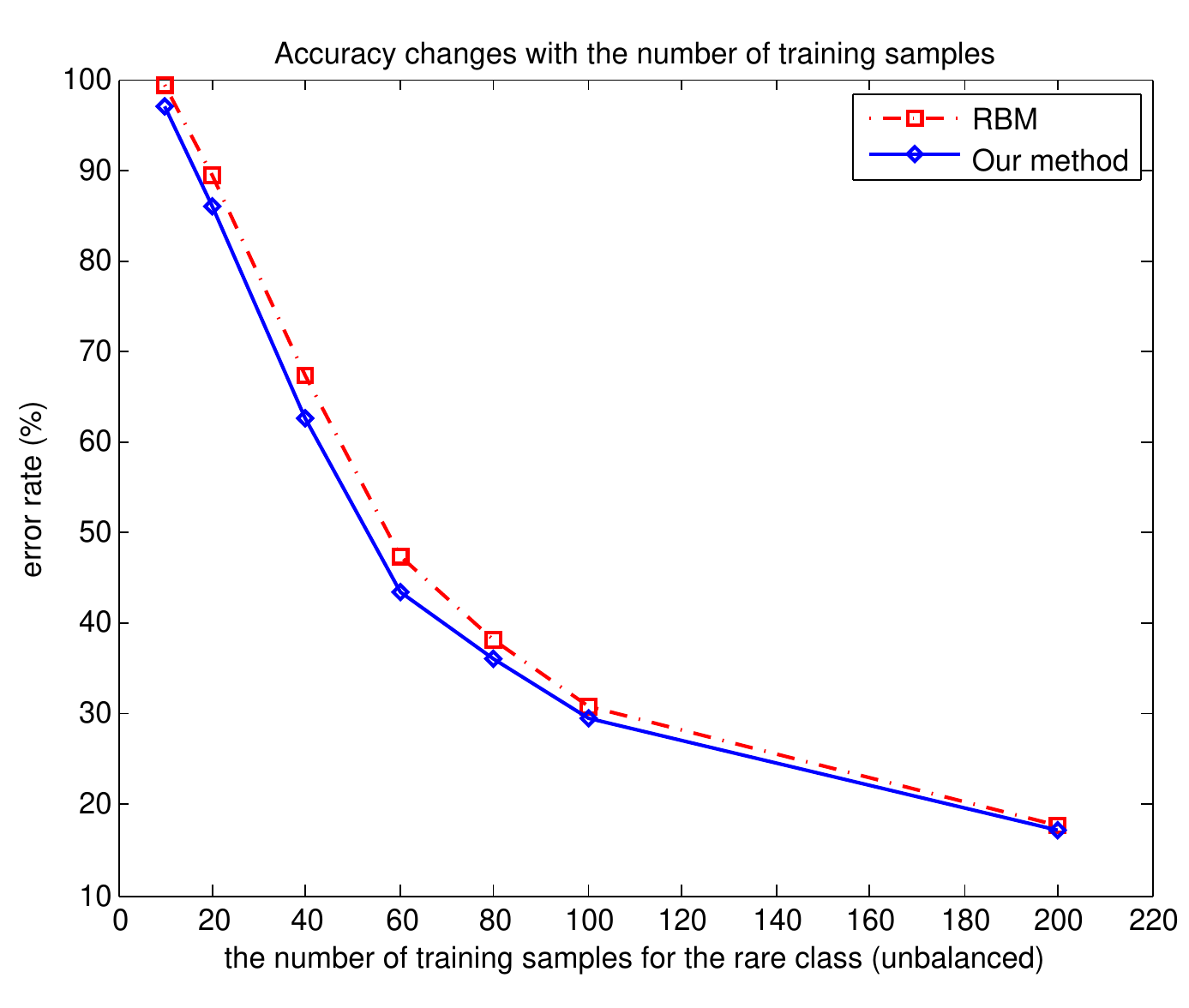}\\
(a) & (b)
\end{tabular}
\caption{(a) How the error rate changes with the balanced training samples; (b) the average error rate on the rare classes, which sample a few examples for each rare class, while keeping other classes each with 500 samples respectively. It shows that our method works better than RBM with few training cases, and yield higher accuracy on the rare classes.}
\label{fig:hier_mnist}
\end{figure*}

To further indicate whether our method is helpful or not with few training samples and how it performs on rare classes, we tested our model on the balanced and unbalanced cases. For the balanced case (each class was sampled equally), we random sampled from 1000 to 5000 examples respectively and tested on the 1000 samples from testing set. The results in Fig. \ref{fig:hier_mnist}(a) demonstrates that our model works better than RBM. We also tested our method on the rare classes. Basically, we sampled a few examples for each rare class, while keep the other classes with 500 samples respectively. For example, we sample the 10 examples for the class `0', while the rest 9 classes have 500 training examples respectively. Then, we training our model with these samples, and test it on the test cases. Similarly, we did the same testing on classes `1' to `9' respectively. We tested our method for each rare class on the testing set, and show the over average error rate in Fig. \ref{fig:hier_mnist}(b), which clearly demonstrates that our method is much better than RBM on rare classes. %In other words, it is helpful to introduce the hierarchical priors into RBMs.

\noindent {\bf Document Classification} We also evaluated our model on 20 news group dataset for document classification. The 20 news group dataset\footnote{\url{http://people.csail.mit.edu/jrennie/20Newsgroups/}} has 18,846 articles with with 61188 vocabularies, which has been widely used in text categorization and document classification. In the experiment, we tested our model on the version of the 20 news group dataset\footnote{\url{http://www.cs.toronto.edu/~larocheh/public/datasets/20newsgroups/20newsgroups_{train,valid,test}_binary_5000_voc.txt}}, in order to make our results comparable to the current state of the art results. In the experiment, we used the hierarchical prior structure over label shown in Fig. \ref{fig:hrbm} for HHRBM, HRBMh, HRBMs and our model. As for parameter setting, we use CD-1, and set the number of hidden states $n = 2000$, learning rate $\eta = 0.1$ and the maximum epoch equals to 100 for RBM related methods. For HHRBM, we set the number of hidden states to be 1000, 500, 200 and 200 respectively for each layer. As for our method, we set $n = 1500$, $\eta = 0.01$, $C=0.1$ and maximum epoch 200. 

The results of different methods are shown in Table (\ref{tab:tab4}). Once again, our model outperforms the other RBM models, also get better results than SVM and neural network classifiers. HRBMs and corrMNL have bad performance in this dataset. The reason we guess is that HRBMs calculates the classification probability for each class by multiplying the output probabilities along the path from root to the leaf associated to that class. Thus, HRBMs will prefer the high level class for unbalanced hierarchical structure. Note that the hierarchical tree in Fig. \ref{fig:hrbm} is unbalanced structure. For HRBMs, `alt.Atheism', `misc.forsale' and `soc.religon.christian' will have higher probability to be labeled compared to leafs (or classes) in the level 4. corrMNL may have the same problem as HRBMs. Another reason for the low performance is that corrMNL does not consider the parameter redundancy problem between adjacent layers as in our model. 

We also evaluate how the regularization term influences the performance. We set $C=0$ to remove the orthogonal restriction, and get accuracy $30.1\%$ in Table (\ref{tab:tab4}), which is significant lower than the result with orthogonal restriction. Hence, it demonstrates that it is useful to introduce orthogonal restriction to the correlated hierarchical prior. 

%\begin{figure}[h!]
%\centering
%\includegraphics[trim = 3mm 1mm 1mm 2mm, clip, width=8.5cm]{hier_20news.pdf}
%\caption{The hierarchical structure from 20 news group dataset. The root (or the first level) cover documents from all categories, while the leaf level indicates labels where documents attached to.}
%\label{fig:hier_20news_prior}
%\end{figure}

\begin{table}[t!]
\centering
\resizebox{0.7\columnwidth}{!}{
\begin{tabular}{lr}
\hline
Model & Error rate (\%) \\
\hline
RBM & 24.9\\ %RBM ($\eta = 0.0005$, $n = 1000$) & 24.9\\
DRBM \cite{Larochelle12} & 27.6 \\% DRBM ($\eta = 0.0005$, $n = 50$) & 27.6\\
RBM + NNet \cite{Larochelle12} & 26.8\\
HDRBM \cite{Larochelle12} & 23.8 \\
HRBMh ($\eta = 0.1$, $n = 2000$) & 30.6\\
HRBMs ($\eta = 0.1$, $n = 2000$) & 63.7\\
HHRBM ($\eta = 0.1$, $n = 1000, 500, 200$ \textrm{and} $200$) & 32.0\\
Ours ($\eta = 0.01$, $n = 1500$ and $C = 0 $) &  30.1 \\
Ours ($\eta = 0.01$, $n = 1500$ and $C = 0.1$) & {\bf 23.6} \\
\hline
MNL & 30.8\\
corrMNL \cite{Shahbaba2007} & 79.3 \\
SVM \cite{Larochelle12} & 32.8 \\
NNet \cite{Larochelle12} & 28.2 \\
\hline
%  20news & 30.8 & - & 30.6 & 0.026 & 32.2 & 24.9 & 23.9\\
\end{tabular}}
\caption{The experimental comparison on 20 news group dataset. We compare the performances between our method and other RBM models. It demonstrates that our method with hierarchical prior over labels can improve recognition accuracy. }
\label{tab:tab4}
\end{table}

% ----------------------------------------------------------------------------------------------
\section{Related work}
The hierarchical structure is organized according to the similarity of classes. Two classes are considered similar if it is difficult to distinguish one from the other on the basis of their representation. The similarity of classes increases as we descend the hierarchy. Thus, the hierarchical prior over categories provides semantic meaning and valuable information among different classes; and thus to some extent it can assist classification problems in hand \cite{Shahbaba2007,Xiao11,Babbar13}. Much work has extensively been done in the past years to exploit hierarchical prior over labels for classification problem,  such as document categorization \cite{Koller97,McCallum98,Weigend99,Dumais00,Cai04} and object recognition \cite{Marszalek07a}. %In general, the hierarchical prior over class labels can provide correlations different categories and assist classification problems with better results. %This hierarchy is used for reasoning and to organize and train the binary SVM detectors. The trained hierarchic classifier can be used to efficiently recognize a large number of object categories. 
Two most popular approaches to leverage hierarchical prior can be categorized below. The first approach classifies each node recursively, by choosing the label of which the associated vector has the largest output score among its siblings till to a leaf node. An variant way is to compute the conditional probability for each class at each level, and then multiply these probabilities along every branch to compute the final assignment probability for each class. Xiao et al. introduced a hierarchical classification method with orthogonal transfer \cite{Xiao11}, which requires the parameters of children nodes are orthogonal to those of its parents as much as possible. Another example is the nested multinomial logit model \cite{Shahbaba2007}, in which the nested classification model for each node is statistically independent, conditioned on its parent in the upper levels. One weakness of this strategy for hierarchical classification is that errors will propagate from parents to children, if any misclassification happened in the top level. 
The other methodology for hierarchical classification prefers to use the sum of parameters along the tree for classifying cases ended at leaf nodes. Cai and Hoffmann \cite{Cai04} proposed a hierarchical larger margin multi-class SVM with tree-induced loss functions. Similarly, Dekel et al. in \cite{Dekel04} improved \cite{Cai04} into an online version for hierarchical classification. Recently, Shahbaba et al. proposed a correlated multinomial logit model (corrMNL) \cite{Shahbaba2007}, whose regression coefficients for each leaf node are represented by the sum of parameters on all the branches leading to that class. 

Apart from the two approaches mentioned above, there are also other methods proposed in the past. Dumais and Chen trained different classifiers kind of layer by layer by exploring the hierarchical structure \cite{Dumais00}. Cesa-Bianchi et al. combined Bayesian inference with the probabilities output from SVM classifiers in \cite{Cesa-Bianchi06} for hierarchical classification. Similarly, Gopal et al. \cite{Gopal12}  used Bayesian approach (with variational inference) with hierarchical prior for classification problems. 

%To the best of our knowledge, no work until now has incorporated hierarchical prior into RBM framework. In this paper, we propose an classification restricted Boltzmann machine with hierarchical correlated prior. %The taxonomy structure arranges similar classes in the lower level of the tree, and diversified classes in the higher level. 
%Basically, we divide the classification RBM into traditional RBM (feature learning or representation) and multi-class logistic regression, and introduce the hierarchical prior over categories into the logistic prediction part in RBM. Thus, our model can capitalize the valuable information endowed with the taxonomy. Moreover, we also force orthogonal constraints between parameters along children-parent pairs in the hierarchy. %To the best of our knowledge, this is the first method, which adds hierarchical prior over RBM model. 
%% -------------------------------------------------------------------------------
%\section{Discussion}
%
%Our method introduces the hierarchical correlation and also orthogonal regularization between parameters. We can think our method is a kind of mixture of corrMNL \cite{Shahbaba2007} and the orthogonal SVM model \cite{Xiao11}. However, our model inherits the advantage of RBM, which can learn the hidden representation for better classification \cite{Hinton06b,Larochelle12}, compared to the multinomial logit \cite{Shahbaba2007} and hierarchical SVM \cite{Dekel04,Xiao11}. Moreover, we only have a single RBM in our model, while there are multiple SVMs in the orthogonal hierarchical SVM \cite{Xiao11}.

\section{Conclusion}
We consider restricted Boltzmann machines (RBM) for classification problems, with prior knowledge of sharing information among classes in a hierarchy. Basically, our model decompose classification RBM into traditional RBM for representation learning and multi-class logistic model for classification, and then introduce hierarchical prior over multi-class logistic model. In order to reduce the redundancy between node parameters, we also introduce orthogonal restrictions in our objective function. To the best of our knowledge, this is the first paper that incorporates hierarchical prior over RBM framework for classification. We test our method on challenge datasets, and show promising results compared to benchmarks.

\bibliography{crbmbib}
\bibliographystyle{iclr2015}

\end{document}